# Comparative Ellipsis and Variable Binding

Jan Lerner and Manfred Pinkal
University of the Saarland
Department of Computational Linguistics

## 1   Introduction

It is hard to find a class of natural language phenomena which exhibits more complex interactions between syntax and semantics than comparative constructions. Comparatives come with phrasal or clausal complements. In this paper, we focus on phrasal comparatives, which on first inspection look simpler than the clausal ones, but in fact add a degree of complexity. We will consider predicative comparatives as in (1) as well as attributive comparatives, which come in two versions, a "narrow" reading as in (2) and a "wide" reading as in (3) (referred to in the following with "NRA" and "WRA", respectively).

*(1)    George is richer than Bill*

*(2)    George owns a faster car than this BMW*

*(3)    George owns a faster car than Bill*

*(4)    George owns a faster car than Bill  owns (a d fast car)*

(1) suggests a direct interpretation of the comparative adjective as a simple two-place relation between standard entities. In (3), the overt elements of comparison – the comparative complement *Bill* and its matrix clause correlate *George* – are related only indirectly, as the clausal paraphrase (4) indicates. Thus, a correspondingly simple treatment is excluded. We will start by analysing the complex case of WRA constructions, arguing that they require the application of operations which make missing material in the complement available before interpretation takes place. We will propose a division of labour between syntax and semantics in the process of recovery of missing information. In Section 2, we will argue that attributive phrasal comparatives are genuinely elliptic, namely a kind of an ACD (antecedent-contained deletion) construction. Syntactic reconstruction leads us from the phrasal version (3) to an LF representation similar to the clausal variant (4) (without the "comparative deletion" portion set in parentheses in (4)). In Section 3, we will spell out the semantics for comparatives. Part of the semantics is an anaphoric mechanism similar to that of one-anaphora, which works for phrasal and clausal comparatives in the same way, providing the information indicated in parentheses in (4). In Section 4, we will discuss whether the reconstruction analysis should be extended to NRA constructions, and argue to treat them differently, as genuinely phrasal constructions with direct semantic interpretation. Section 5 among other things discusses the semantics of predicative constructions.

## 2   A reconstruction analysis for WRA comparatives

### 2.1   Irene Heim's Direct Analysis

Heim (1985) proposes an interesting alternative to a reconstruction analysis for phrasal comparatives. It saves the idea of a comparison between the overt correlates in WRA constructions, by allowing the property with respect to which the comparison takes places to be more complex than just the degree property expressed by the adjective. She assumes (5), (6) and (7) as the structures underlying the semantic interpretation for sentences (1), (2) and (3), respectively.

(5)    $er\_than(<George, Bill>) ( \lambda x \; \iota d[x \; is \; d \; rich])$

(6)    $\exists y \; [er\_than \; (<y, this\_BMW>) \; (\lambda x \; \iota d[d \; fast \; car \; (x)]) \; \& George \; owns \; y]$

(7)    $er\_than \; (<George, Bill>) \; ( \lambda x \; \iota d[x \; owns \; a \; d \; fast \; car])$

The semantic interpretation of the *er_than* operator is given in (8): It takes a pair of individuals and a function from individuals to degrees, and compares the degrees assigned to the individuals by that function. The resulting truth conditions for the WRA case (3)/(7) are paraphrased in (9).

(8)    $[\![ er\_than <a,b> f ]\!] = 1 \; iff \; f(a) > f(b)$

(9)    The (maximal) degree d such that George owns a d fast car is higher than the (maximal) degree d such that Bill owns a d fast car.

Heim's proposal demonstrates that it is possible to arrive at an interpretation for WRA phrasal comparatives without reconstruction of a clausal complement. However, it is not completely satisfactory for syntactic as well as semantic reasons.

The structures (5)-(7) underlying interpretation are semantically motivated. They are quite remote from what could be considered as a natural syntactic analysis for (1) - (3), and it requires several non-standard assumptions to obtain them. The *er_than* operator heading the whole structure corresponds to a surface element which occurs as an inflection morpheme of the deeply embedded adjective. The status of the operations moving complement and correlate NP out of their different surface positions into the immediate domain of the comparative operator is unclear. The iota operator binds the degree variable across an NP and AP boundary.

Semantically, Heim's direct analysis yields an appropriate interpretation only for those attributive constructions occurring with a plain indefinite comparative NP, like (3). The symmetry in the treatment of complement and correlate leads to inadequate truth conditions for cases (10) - (13). The English examples (10) and (12) seem to be somewhat marginal. The corresponding German sentences (11) and (13) are perfectly acceptable, as are all kinds of attributive comparatives that come with a symmetric determiner.

(10)   George owns at least two faster cars than Bill

(11)   George besitzt mindestens zwei schnellere Wagen als Bill

(12)   George owns at most one faster car than Bill

(13)   George besitzt höchstens einen schnelleren Wagen als Bill

The intuitions about (11) are that it is true if George owns at least two cars which are faster than any car of Bill's. According to Heim, we have to assume (14)/(15) as input structure for the interpretation of (10)/(11), and obtain the truth conditions in (16), which are clearly inadequate.

(14)   *er_than (<George, Bill>) (λx ιd[x besitzt wenigstens zwei d schnelle Wagen])*

(15)   *er_than (<George, Bill>) (λx ιd[x owns at least two d fast cars])*

(16)   The (maximal) degree d such that George owns at least two d- fast cars is higher than the (maximal) degree d such that Bill owns at least two d- fast cars.

The direct analysis also leads to wrong predictions about anaphoric accessibility properties of WRA comparatives.

(17)   *George owns a faster car than Bill. It (= George´s car/ * Bill´s car) is a BMW*

As (17) illustrates, George's car is accessible for anaphoric reference from outside, Bill's car is not. This asymmetry is clearly a problem for a theory which is based on the assumption that the function of the indefinite NP is to determine a degree property which then is applied to complement and correlate in precisely the same way. In both kinds of counter-examples, it seems to be the symmetry in Heim's analysis which is the main obstacle to an appropriate interpretation. In the following, we will propose a reconstruction analysis, which takes the different semantic status of correlate and complement into account and gets along without special assumptions for the syntax of comparative constructions.

## 2.2   The Syntactic Position of the Comparative Complement

The topicalization tests in (18)-(25) show that the comparative complement forms a constituent with the comparative AP/NP. This holds for English as well as for German, and leads us to assume the over-all syntactic structures indicated in (26) and (27), respectively.

(18)   *Richer than Bill is George indeed*
(19)   *Reicher als Bill ist George tatsächlich*

(20)   *? Richer is George than Bill*
(21)   *? Reicher ist George als Bill*

*(22)    A faster car than Bill, George owns indeed*
*(23)    Einen schnelleren Wagen als Bill besitzt George tatsächlich*

*(24)    ?A faster car, George owns than Bill*
*(25)    ? Einen schnelleren Wagen besitzt George als Bill*

*(26)    George is [$_{AP}$ richer than Bill]*

*(27)    George owns [$_{NP}$ a faster car than Bill]*

In (27), there are two possible options for the syntactic position of the *than* phrase: adjunction to NP or adjunction to N'. There is a bundle of related phenomena that strongly suggest the adjunction-to-NP version. First, look at examples (28) (=(3)) - (30), which demonstrate the so-called indefiniteness effect.

*(28)    George owns a faster car than Bill*

*(29)    *George owns every faster car than Bill*

*(30)    *George owns the faster car than Bill*

The comparative complement interacts with the determiner of the comparative NP, rendering, e.g., definite and universally quantified cases unacceptable. If we take the comparative complement to be an N' modifier, a straightforward semantic interpretation would make the complement contribute to the restriction of the respective determiners ("car which is faster than any car that Bill owns"), and there it should work as good for *every* and *the* as it does for the indefinite. Furthermore, the definiteness effect is dependent on the explicit occurrence of a complement phrase, as shown by the acceptability of (31)-(33).

*(31)    George owns a faster car*

*(32)    George owns every faster car*

*(33)    George owns the faster car*

Finally, we observe that there is no definiteness effect either in cases where an overt complement exists, but is not adjoined to the comparative NP. Adjunction to NP is impossible in the German examples with pre-nominal complement (34)-(36), and is implausible at least in the English post-nominal comparatives as in (37)-(39).

*(34)    ein mehr als 100 m hohes Gebäude*
*       a more than 100 m high building*

*(35)    das mehr als 100 m hohe Gebäude*
*       the more than 100 m high building*

*(36)    jedes mehr als 100 m hohe Gebäude*
*       every more than 100 m high building*

*(37)    a building higher than the ET*
*(38)    the building higher than the ET*

(39)    *every building higher than the ET*

It seems to be crucial for the definiteness effect to come about that the determiner occurs in the scope of the comparative phrase. Therefore we assume (40) as constituent structure for the attributive comparative construction (3).

(40)    *George owns [$_{NP}$ [$_{NP}$ a faster car] than Bill]*

(41)    *George is [$_{AP}$ [$_{AP}$ richer] than Bill]*

We may adopt the corresponding analysis (41) for the predicative case, for reasons of parallelism, although there is no independent evidence for that.

## 2.3  Phrasal Comparatives as ACD Constructions

The syntactic considerations of the last section lead to a configuration for the attributive construction which perfectly corresponds to the antecedent-contained deletion (ACD) cases discussed in May (1985) and Fiengo/May (1991); cf. (42) and (43).

(42)    *John read [$_{NP}$ [$_{NP}$ every book] Bill did e]*

(43)    *George owns [$_{NP}$ [$_{NP}$ a faster car] than Bill e]*

In ACD constructions like (42), the elided portion of the elliptic construction is itself part of the antecedent. Therefore, naive application of a copying operation to the antecedent would lead to an infinite regress. The way out is provided by the operation of Quantifier Raising, which allows removal of the constituent containing the elided part before copying.

We will outline how the ACD treatment proposed by Fiengo/May (1993) can be adapted to the case of the attributive comparative construction. First, we QR the comparative NP in (43). The result, (44), is transformed to (45), by applying QR another time to *George*, in order to move the correlate out of the IP of the matrix clause. Next, we copy the IP, and obtain (46). We take the copy of the subject variable $t_1$ to receive its index from the complement NP *Bill*, which is a standard assumption, and also in accordance with the theory of Fiengo/May (1993), since $t_2$ is an i-copy of $t_1$.

(44)    *[$_{NP}$ [$_{NP}$ a faster car] than Bill e]$_i$ [George owns $t_i$]*

(45)    *[$_{NP}$ [$_{NP}$ a faster car] than Bill e]$_i$ [George$_1$ [$_{IP}$ $t_1$ owns $t_i$]]*

(46)    *[$_{NP}$ [$_{NP}$ a faster car] than Bill$_2$ [$_{IP}$ $t_2$ owns $t_?$] ]$_i$ [George$_1$ $_{IP}$ $t_1$ owns $t_i$]]*

The second variable in the copied IP of (46) has not found a binder yet. Chomsky (1977) proposes an analysis of clausal comparatives, where the empty noun phrase of the complement clause is bound by an equally empty wh-operator. His analysis for (47) is indicated in (48).

(47)    *John wrote more books than Bill read*
(48)    *John wrote more books than [WH$_1$ [Bill read $t_1$]]*

The ACD analysis of phrasal comparatives has lead us to a level of reconstruction which corresponds to the clausal construction, providing the complement with all material necessary for its interpretation except the degree-containing phrase itself. This suggests adopting the structural assumptions made by Chomsky for clausal constructions to our ellipsis case, i.e., to assume the presence of an operator in the comparative complement which binds the copied NP variable. The resulting structure for sentence (3) is given in (49).

(49)     $[_{IP}[_{NP} [_{NP}$ a faster car] than $[_{CP} WH_j\ C\ [_{IP}\ Bill_2\ [_{IP}\ t_2\ owns\ t_j]]]]_i$
                    $[George_1[_{IP}\ t_1\ owns\ t_i]]]$

Since we take the operator to be present from the beginning as part of the complement construction, the question remains as to how the co-indexing with the copied variable is achieved. A simple answer would be that the operator needs an argument to bind and takes just the unbound variable it finds in its scope. If we adopt the theory of Fiengo and May, we must go into slightly greater detail. The structural descriptions are different in the dependencies <(WH,t),j, <WH,C,NP,NP,NP,V,NP>> and < (a faster car than Bill owns,t),i, <NP,NP,NP,V,NP>>. Therefore we cannot come up with an i-copy for $t_i$, as we did for $t_1$, but have to assume a strict copy of $t_j$ instead, i.e., a copy bearing the same index i. Thus, the question whether WH can or cannot bind the copied variable properly depends on the value of its index j. Now, as an empty category, WH must have an antecedent, which is the NP *a faster car*. In turn, this NP inherits its index from the complex NP which it is part of. Since this index is i, $t_{j(i=j)}$ is properly bound and a correct copy of $t_i$.

This concludes the syntactic part of our story. Let us see what we have achieved so far. By adopting an ACD type reconstruction analysis, we have moved a considerable step forward to a solution of the problems listed in 2.1.

- By QRing the comparative NP, its determiner is removed from the domain of reconstruction. Thus it cannot induce an inappropriate interpretation of the complement (Examples (10)-(13)) .

- The asymmmetric syntactic structure favors (or at least, does not exclude) an asymmetric treatment of anaphoric accessibility (Example (17)).

- The reconstructed structure relates to a well-motivated surface constituent structure in a standard way. In particular, all variables are syntactically motivated and properly bound.

There is a big open problem, however. If we look at our resulting structure (49), which tells a story about the binding of NP variables and does not make any reference to degrees, it seems that we might have made a step away from the required semantics. We need to clarify how the appropriate semantic interpretation is achieved on the basis of (49). We will approach this task in the following section.

## 3 Semantic Interpretation for WRA Comparatives

### 3.1 Interpretation of the Comparative Complement

We are faced with a mismatch concerning the status of the gap in the (reconstructed) complement clause: according to the syntactic analysis of the last section, it is a NP variable ranging over standard individuals bound by an empty operator inside the complement clause. According to the requirements of semantic analysis, however, the comparative complement should denote a degree term, in fact, a universal degree quantifier, as has been argued for in detail in Lerner/Pinkal (1992). The interpretation of the complement of (3) should be something like (50), which is paraphrasable as (51), rather than "(faster than) what Bill owns", which is the presumable result of making straightforward sense out of the reconstructed complement *than WH$_i$ Bill owns t$_i$* .

(50)  $\lambda D´\forall d\ [\exists y[fast'(car')(y,d) \wedge own'(b^*,y)] \rightarrow D'(d)]$

(51)  "(faster than) every degree such that Bill owns a d fast car"

In order to achieve a sensible interpretation of the complement, semantic analysis must relate the variable introduced by syntactic reconstruction, which ranges over standard individuals, to a variable ranging over degrees. In order to do this appropriately, it must make reference to the relation encoded in the N' of the comparative NP. Also, it must take care that the degree variable is bound by an operator with an appropriate semantics. According to what has been said before, we will assume a universal degree quantifier. Actually, it is of secondary interest here whether it is a universal quantifier, or a definiteness or maximum operator, but it is of primary importance that it is a <u>degree</u> operator (for a discussion of this problem see also Moltmann (1992)).

In order to make the content of the N' available, we assume an anaphoric mechanim which is similar to One Anaphora. This view is supported by the fact that German clausal comparatives like (52) employ an overt anaphoric element that is morphologically identical to the pronoun used in One Anaphora.

(52)  *George besitzt einen schnelleren Wagen als Bill einen besitzt*
       *George owns a faster car than Bill one owns*

Also, as mentioned already in the initial discussion of WRA constructions with *at least two*, *at most one, etc.,* the semantic complement clause counterpart of the comparative NP is always a simple existential quantifier, which fits nicely into the semantics of standard one anaphora.[1] The only specific feature of the comparative anaphora is that it refers to an N' content containing an open degree position ("d fast car"). The anaphoric element, which we will call **P$_0$** in the following, is actually a relation between individuals and degrees, rather than a one-place predicate. The semantic components which the comparative complement construction contributes - an anaphoric component similar to One Anaphora, and a degree operator in terms of a universal determiner - are spelled out in (53) and (54) in type-theoretic notation. The **P$_0$** variable in (53) would have to be instantiated to fast'(car') in our standard example.

(53)     $\lambda Q \exists y[\mathbf{P_0}(y,d) \wedge Q(y)]$

(54)     $\lambda D \lambda D' \forall d\ [D(d) \rightarrow D´(d)]$

These two ingredients are definitely required – or variants or them which serve the same purpose. The question is at which place they come into play. One option would be to interpret the empty direct object position in the comparative complement by (53), and the operator assumed in the syntactic analysis by (54). Binding x in own(b*,x) by (53) gives (55), binding d in (55) by (54) gives (56), and by instantiating the anaphoric variable in (56) with the predicate "fast'(car')" it refers to, we obtain (50), the intended interpretation for the comparative.

(55)     $\exists y[\mathbf{P_0}(y,d) \wedge own(b^*,y)]$

(56)     $\lambda D' \forall d\ [\exists y[\mathbf{P_0}(y,d) \wedge own(b^*,y)] \rightarrow D´(d)]$

The problem with this analysis is that the semantics tells a different story of the binding relation between operator and variable than the syntax does. The standard NP binding operator has to be re-interpreted in some way or the other as a degree binding operator. Another option might be to have the anaphoric indefinite NP introduced as above, but then raise it to a position where the degree determiner can bind its degree variable position, before it is applied to own(b*,x). Semantically, the combination of (54) and (53) is no problem as soon as (53) is appropriately abstracted over; generalized functional application (see below) gives us (57), which in turn can be applied to own(b*, x) to yield the intended analysis.

(57)     $\lambda Q \lambda D' \forall d\ [\exists y[\mathbf{P_0}(y,d) \wedge Q(y)] \rightarrow D'(d)]$

Syntactically, the solution has the advantage, that the NP variable in the syntax of the complement translates to a standard individual variable in the semantics and does not need any special treatment. However, it requires the assumption of an additional empty position and of a "hidden" degree variable which serves as the argument of the operator. Therefore, we choose the syntactically simplest solution: We do not extend the syntax by any additional element or operation, but attach both the anaphoric and the quantificational aspect to the empty WH operator. Thus, we burden the WH operator with a lot of semantic information. This might look like a hack, but consider that it is just the information that is induced invariably by any comparative complement construction (at least in the WRA cases). The alternatives that came to our minds maybe look simpler since they distribute the semantic material. However, they rather increase syntactic complexity.

We would like to conclude this section by pointing to a desirable side effect of our analysis. Closer inspection of the syntactic analysis given to the complement in (49), repeated here as (58), shows that there is a second way of interpretation: according to May´s Scope Principle (May 1985), the raised NP *Bill* may take scope over the WH-operator, since both occur in the same c-command domain.

(58)     [$_{CP}$WH$_j$ C [$_{IP}$Bill$_2$ [$_{IP}$ t$_2$ owns t$_j$]]

In the case of our example, the difference is truth-conditionally irrelevant, since the involved NP is a proper noun. Cases where genuine quantifiers and other logical operators occur in the comparative complement show that both scoping variants are needed.

(59)  *George has a faster car than any policeman*

(60)  $\lambda D' \forall d\ [\exists y[\mathbf{P_0}(y,d) \land \exists x\ [policeman'(x) \land has'(x,y)] \rightarrow D'(d)]]$

(61)  *George has a faster car than every policeman*

(62)  $\lambda D' \forall d\ [\exists y[\mathbf{P_0}(y,d) \land \forall x\ [policeman'(x) \rightarrow has'(x,y)] \rightarrow D'(d)]]$

*any policeman* in (59) must take narrow scope under the WH operator, since it requires the downward entailing context provided by the universal degree quantifier, which correctly results in the interpretation (60) (with $\mathbf{P_0}$ anaphorically relating to fast'(car')). *every policeman* in in (61), on the other hand, can and must take scope over the WH operator in order to yield an intuitively appropriate interpretation: narrow scope interpretation along the lines of (60) yields (62), which considers only the fastness degrees of those cars owned by every policeman at the same time. The correct reading is brought about by first applying the WH operator (57) to $\lambda y\ has'(x,y)$, resulting in (63), and next quantifying $\lambda G \forall x\ [policeman'(x) \rightarrow G(x)]$ into the resulting expression, from outside.

(63)  $\lambda D' \forall d\ [\exists y[\mathbf{P_0}(y,d) \land has'(x,y)] \rightarrow D'(d)]$

The problem is that standard functional application cannot be used in this case, since (63) does not have the appropriate type: It is not a formula, but a degree quantifier, or, in other words, a formula lacking a degree predicate. For several independent reasons, we found it useful to employ a liberalized version of Functional Application here. In this case, Functional Composition (FC) would do as well, but other cases of composition suggest an operation that, unlike FC, binds the elements on the lambda list of the argument term from outside. We call this operation GFA ("Generalized Functional Application") and use the infix operator •. GFA is similar to Functional Composition in that it passes up unsaturated "lambda requirements" of the argument to the representation of the mother node. GFA can be defined in terms of plain functional application, and thus is a logically harmless extension of the type-theoretic representation language[2].

The GFA in (64) gives us (65), the intuitively correct reading of (61):

(64)  $\lambda G \forall x\ [policeman'(x) \rightarrow G(x)]$
       • $\lambda x\ \lambda D' \forall d\ [\exists y[\mathbf{P_0}(y,d) \land has'(x,y)] \rightarrow D'(d)]$

(65)  $\lambda D'\ \forall x\ [policeman'(x) \rightarrow \forall d\ [\exists y[\mathbf{P_0}(y,d) \land has'(x,y)] \rightarrow D'(d)]]$

For (66), the application of May's scope principle predicts an ambiguity between readings (67) and (68), in accordance with intuitions.

*(66)    George owns a faster car than Bill or Richard*

*(67)    George owns a faster car than Bill or George owns a faster car than Richard*

*(68)    George owns a faster car than both Bill and Richard*

The scope interaction between comparatives and logical operators in their complements has been observed before. Unlike other theories, which have to assume a non-standard quantifier raising operation moving the NP out of the complement to the top of the matrix clause (cf. v. Stechow 1984), our account produces these results directly without any additional stipulation, given the operation of Generalized Functional Application.

### 3.2   Interpretation of the Comparative NP

So far, we have only specified the semantics of the comparative complement. As a prerequisite, we assumed for the adjective *fast* a semantics, which is explicitly given in (69). It is a predicate modifier with an additional open degree position, which has to be bound in some way or the other.

(69)    *fast*    ⇒    $\lambda d \lambda Q \lambda x\ \text{fast}'(Q)(x,d)$

The comparative complement must be involved in the binding of the degree variable. The complement term is a degree quantifier, but it should not bind the degree position of the adjective directly. The binding relation must be mediated by the comparison relation, a "greater than" relation between degrees: the comparative assigns its external argument a degree greater than whatever the complement specifies. Technically, the comparative operator binds the degree argument of the matrix sentence adjective existentially, and relates it to the degree term specified by the complement. The question is which part of the construction repeated in (70) should be regarded as the syntactic realization of the comparative operator.

*(70)    [$_{NP}$ [$_{NP}$ a [$_{AP}$ faster] car] [$_{PPt}$ han [WH$_j$ [$_{IP}$ Bill$_2$ [$_{IP}$ t2 owns t$_j$]]]]]*

There are two theoretical options: the comparative morpheme *er* or the particle *than*, which has not been assigned a semantic function so far. The latter case however would only cover the comparative constructions with explicit complement. Therefore, we take *than* to be a semantically empty element and interpret the comparative morpheme as in (71), where $\mathbb{P}$ is a variable ranging over degree quantifiers.

(71)    *er*    ⇒    $\lambda D \lambda \mathbb{P}[\exists d'\ [\ \mathbb{P}(\lambda d[d' > d])\ \wedge\ D(d')]]$

(71)•(69) gives (72), application of (72) to the head noun of the NP gives (73). GFA of the standard representation of the indefinite article to (73) is (74), and application of (74) to the degree determiner (56) denoted by the comparative complement results in (75), the representation for the comparative NP (70).

(72)   *faster*  ⇒
  λDλℙ[∃d' [ ℙ(λd[d'>d]) ∧ D(d')]] • ( λdλQλx fast'(Q)(x,d))
    ⇔   λQλxλℙ ∃d' [ ℙ(λd[d'´>d]) ∧ fast' (Q)(x,d')]

(73)   *faster car*   ⇒ λxλℙ ∃d' [ ℙ(λd[d'>d]) ∧ fast' (car')(x,d')]

(74)   *a faster car*  ⇒ λℙλQ∃x ∃d' [ ℙ(λd[d'>d]) ∧ fast' (car')(x,d') ∧ Q(x)]

(75)   *a faster car than Bill (owns)*  ⇒
  λQ∃x ∃d'[∀d [∃y[P$_0$(y,d) ∧ own'(b*,y)] → d'>d]
            ∧ fast' (car')(x,d') ∧ Q(x)]

The semantic representation (76) for our example sentence (3) results from quantifying (75) into own'(g*, y), and instantiating **P$_0$** with fast'(car').

(76)   *George owns a faster car than Bill*  ⇒
  ∃x ∃d' [ ∀d [∃y [fast'(car')(y,d) ∧ own'(b*,y)] → d'>d]
              ∧ fast'(car')(x,d') ∧ own'(g*,x)]

We come back to the semantic problems raised by the direct analysis of Heim (1985). First, the two underlined existential quantifiers in (76), which correspond to George's and Bill's car, respectively, have different status: the first one is a top-level existential, the second is dependent on a universal degree quantifier. This solves the problem of asymmetric anaphoric accessibility mentioned in 2.1. Second, the analysis models the truth-conditional asymmetry in attributive constructions with cardinality specifications, as the representation (77) of example sentence (10)/(11) shows (for the sake of simplicity, we used the English variant here although it is less acceptable).

(77)   *George owns at least two faster cars than Bill*  ⇒
  ∃²**x** ∃d' [ ∀d [∃**y**[fast'(car')(y,d) ∧ own´(b*,y)] → d'>d]
               ∧ fast'(car')(x,d') ∧ own'(g*,x)]

## 4  A direct analysis for NRA comparatives

We proposed a treatment for WRA phrasal comparative constructions, and it seems plausible to extend this analysis to the narrow reading cases like (78) (=(2)), to obtain an input structure to semantic interpretation which corresponds to the clausal paraphrase (79).

*(78)   George owns a faster car than this BMW*

*(79)   George owns a faster car than this BMW is*

Closer inspection however shows that a reconstruction analysis of NRA comparatives poses several problems. First, the subject of a reconstructed IP should be nominative. But as the German examples (80) and (81) show, the case of the complement NP covaries with the case of the comparative NP.

(80)   George besitzt einen schnelleren Wagen als diesen BMW
       George owns a faster car (acc) than this BMW (acc)

(81)   George fährt mit einem schnelleren Wagen als diesem BMW
       George drives with a faster car (dat) than this BMW (dat)

A second problem concerns the status of the reconstructed copula: If NRA constructions are analysed as cases of ellipsis, it is completely open where the tensed form of *be* comes from, since it does not occur in any part of the antecedent. On the other hand, ellipsis would lack its proper function – making non-local linguistic information available – since the resulting structure is completely predictable from the local configuration. Third, there is a strong constraint on the occurrence of different noun-phrase types in the comparative complement: Unlike WRA constructions (and predicative constructions), NRA constructions admit only referential NPs, as (82) demonstrates. This is difficult to explain, if the complement NP of a WRA comparative is taken to be the subject of an underlying complement clause.

(82)   ?George owns a faster car than every BMW

For these reasons, we propose to analyse NRA comparatives in terms of a direct, reconstruction-free interpretation. We assume the syntactic analysis in (83), where the comparative complement is a small clause with the overt complement NP as subject and the WH operator as predicate.

(83) George owns [$_{NP}$ [$_{NP}$ a faster car][$_{PP}$ than [$_{SC}$ [WH [$_{NP}$ Bill]]]]]

The translation of the WH operator is given in (84), where $\mathbb{P}$ is a variable of type <<e,t>,t> and $P_0$ is an anaphoric relational element as in the WRA case.

(84) $\lambda \mathbb{P}[\lambda D [\forall d[ \mathbb{P}(\lambda x[P_0(x,d) ] \to D(d)]]]$

Application of (84) to *this BMW* gives (85), and by instantiating $P_0$ in the appropriate way and carrying out the further interpretation steps described in Section 3.2, we arrive at (86) as the representation of (2)/(78).

(85)   $\lambda D [\forall d[P_0(bmw^*,d) \to D(d)]]$

(86)   $\exists x \exists d' [ \forall d[fast'(car')(bmw^*,d) \to d'>d]$
       $\land\ fast'(car')(x,d') \land\ own'(g^*,x)]$

We have obtained this result without reconstruction. Thus no incorrect assumptions had to be made about the case of the complement or the structure of a reconstruction domain. Note also that Mays scope principle does not apply in the structure (83), which excludes the derivation of a distributive reading of the complement NP as in the WRA case. Also, it is impossible to raise the complement NP out of the comparative NP and adjoin it to the matrix clause IP. The complement NP is captured within the scope of the WH operator, which explains the markedness of the examples in (82).

# 5  Concluding Remarks

In this paper, we discussed the question whether phrasal comparatives should be given a direct interpretation, or require an analysis as elliptic constructions, and answered it with Yes and No. The most adequate analysis of wide reading attributive comparatives seems to be the treatment as ellipsis, where a direct (but asymmetric) analysis fits the data for narrow scope attributive comparatives. The question whether it is a syntactic or a semantic process which provides the missing linguistic material in the complement of WRA comparatives has also been given a complex answer: Access to the linguistic context takes place by a combination of a reconstruction operation and a mechanism of anaphoric reference. It is an advantage of the analysis that it makes only few and straightforward syntactic assumptions. That it can do so is in part due to the availability of the operation of Generalized Functional Application, on the side of semantics, which allows us to model the semantic composition process in a more flexible and, to our minds, more natural way.

A couple of open questions are left. One of them concerns the semantics of predicative phrasal comparative constructions, which we have not considered so far.

*(87)   George is richer than Bill*

Intuitively, examples like (87) (= (1)) are as simple or yet simpler than NRA comparatives. Therefore it is tempting to try just to extend the NRA analysis to the predicative case. There is an additional syntactic argument against a reconstruction treatment. If predicative comparatives are analyzed as ellipsis, they are ACD cases as well and require a raising operation which removes the elided portion from the reconstruction domain. Since the phrase to be raised is an AP, this would constitute a new type of operation which has no independent motivation, to our knowledge. Actually, the direct analysis proposed for NRA constructions can be applied without any changes to (87), (88) being the assumed syntactic structure and (89) the interpretation of the complement.

*(88)   George is [$_{AP}$ [$_{AP}$ richer][$_{PP}$ than [$_{SC}$ [WH [$_{NP}$ Bill]]]]]*

(89)   $\lambda D\ [\forall d[rich'(b^*,d) \rightarrow D(d)]]$

*(90)   George is richer than every professor*

Example (90) shows that attributive constructions do not impose the same constraints on the complement NPs as NRA comparatives do. This can be explained by the fact that the complement NP is not captured under an NP node and therefore can be QRed to the IP node of the sentence. However, sentences like (91) pose a serious problem.

*(91)   George is richer than last year*

We cannot see how our direct analysis could be extended to cases like this. But if we have to analyze (91) as ACD cases, it seems we have not gained a lot by giving (88) a direct interpretation. We have not found a good answer to this problem yet.

The second problem we want to address is connected with the indefiniteness effect. In Section 2.2, we took the constraints on determiners in the comparative NP as evidence for a semantic interaction between complement and the comparative NP as a whole, which was one motivation for an Adjunction-to-NP analysis. However, we have not said so far how this interaction induces the effect. Our tentative explanation goes as follows: The WH operator in the complement is an empty anaphoric element. Therefore, it must be controlled by an antecedent that c-commands it, and the antecedent must be semantically appropriate. These conditions are vacuously satisfied in complement-free-constructions like (94). They are also satisfied in (95) and (96), since in these cases the adjective c-commands the WH.

*(92) (= (3))*   *George owns a faster car than Bill*

*(93) (= (29))*   *\*George owns every faster car than Bill*

*(94) (= (32))*   *George owns every faster car*

*(95) (= (36))*   *jedes mehr als 100 m hohe Gebäude*
              *every more than 100 m high building*

*(96) (= (39))*   *every building higher than the ET*

The situation is more difficult in the standard type of WRA construction. Here, the adjective+CN phrase cannot serve as a controller if we assume an adjunction to NP structure, since in this case the complement is not in its c-command domain. Thus, both (92) and (93) should be ungrammatical, according to a strict reading of our conditions. We may argue, however, according to the lines of DRT and File Change Semantics that the indefinite article does not really add to the semantics, and thus the indefinite NP is semantically very similar to the predicate denoted by the N', or, to put it in a different way, that, different from genuine quantifiers, the indefinite NP is transparent for the predicative part of its content. This would explain why (92) is acceptable in contrast to (93). It would, however, rule out *at most one faster car* along with (93). To explain the acceptability of its German counterpart, we might appeal to the notion of "adjectival character" of an NP, introduced in Higginbotham (1987), and assume that this type of NP is another kind of semantically appropriate antecedent for our WH operator. But then we would have to explain the markedness of the corresponding English occurrences. Although we feel that we are approaching the right line of explanation for the definiteness effect, we must leave the details open.

**Endnotes**

\*     The research reported in this paper was supported by the Deutsche Forschungsgemeinschaft (Project ELAN, grant St 220/4). We thank Irene Heim and Sebastian Millies for helpful comments and suggestions, as well as the participants of SALT and of the Colloquium of the Max-Planck-Forschungsgruppe in Berlin, where versions of this paper have been presented.
1     Furthermore, One Anaphora as a semantic rather than a syntactic process allows flexible access to semantic information irrespective of the way it is syntactically encoded. This makes our analysis compatible with the fact discussed in Bierwisch (1987) that it is not the full semantics of the adjective but rather its dimension-denoting part which is made use of in the interpretation of the complement. E.g., *George is 10 cm shorter than Bill* does not state a difference in the degree of shortness between George and Bill, but a difference in height.

2      GFA of  ϕ to ψ is defined by:   ϕ • ψ = λ$\vec{\sigma}$ ϕ (λ$\vec{v}$ [ψ($\vec{v}$)($\vec{\sigma}$)]). Standard Functional Application is the special case with empty $\vec{\sigma}$. Compare the different effects of GFA and Functional Composition (FC).

FC:    λPP(b*) · λyλxF(x,y)  = λy F(b*,y)

GFA:   λPP(b*) • λyλxF(x,y)  = λx' λPP(b*)(λy'(λyλxF(x,y))(y')(x'))

⇔ λx' λPP(b*)( λy'F(x',y')) ⇔ λx' F(x',b*)

**References**

Bierwisch, M. (1987): Semantik der Graduierung. In *Studia grammatica* XXVI+XXVII, Akademie-Verlag, Berlin.

Chomsky, N. (1977): "On wh movement". In P. Culicover et al. (eds): *Formal Syntax.* Academic Press, New York.

Fiengo, R/ May, R. (1991): Indices and Identity. The MIT Press, Cambridge, Massachusetts.

Heim, I. (1985): Notes on Comparatives and related matters. MS., Austin, Texas.

Higginbotham, J. (1987): Indefiniteness and Predication. In E.J. Reuland/A. G.B. ter Meulen: *The Representation of (In)definiteness.* The MIT Press, Cambridge, Massachusetts.

Lerner, J./Pinkal,M. (1992): Comparatives and Nested Quantification. In M.Stokhof/P. Dekker(eds): Proceedings of the Eighth Amsterdam Colloquium, ILLC, University of Amsterdam.

May, R. (1985): Logical Form. Its Structure and Derivation. MIT Press, Cambridge, Massachusetts.

Moltmann, F.(1992): The Empty Element in Comparatives. NELS 23.

von Stechow, Arnim (1984): Comparing Semantic Theories of Comparison. *Journal of Semanticcs*  3 (1984), 1-77.